\pdfoutput=1

\documentclass[11pt]{article}
\usepackage{multirow}
\usepackage{graphicx}
\usepackage{hyperref}
\usepackage{caption}
\usepackage{subcaption}
\usepackage{amssymb}
\usepackage{pifont}
\newcommand{\cmark}{\ding{51}}%
\newcommand{\xmark}{\ding{55}}%
\usepackage[dvipsnames]{xcolor}

\usepackage[]{naacl2021}

\usepackage{times}
\usepackage{latexsym}

\usepackage[T1]{fontenc}

\usepackage[utf8]{inputenc}

\usepackage{microtype}

\title{A Privacy-Preserving Approach to Extraction of Personal Information through Automatic Annotation and Federated Learning  }

\author{Rajitha Hathurusinghe{$^{1}$}\thanks{\quad Equal contribution.}, Isar Nejadgholi{$^{2}$}\footnotemark[1], Miodrag Bolic{$^3$} \\
  {$^{1,2,3}$}University of Ottawa, Ottawa, Canada\\
  {$^2$}National Research Council Canada, Ottawa, Canada \\
  \footnotesize\texttt{{$^{1,3}$}\{rhath050,Miodrag.Bolic\}@uottawa.ca}\\   \footnotesize\texttt{{$^{2}$}isar.nejadgholi@nrc-cnrc.gc.ca}}

\begin{document}
\maketitle
\begin{abstract}

We curated WikiPII, an automatically labeled dataset composed of Wikipedia biography pages, annotated for personal information extraction. Although automatic annotation can lead to a high degree of label noise, it is an inexpensive process and can generate large volumes of annotated documents. We trained a BERT-based NER model with WikiPII and showed that with an adequately large training dataset, the model can significantly decrease the cost of manual information extraction, despite the high level of label noise. In a similar approach, organizations can leverage text mining techniques to create customized annotated datasets from their historical data without sharing the raw data for human annotation. Also, we explore collaborative training of NER models through federated learning when the annotation is noisy. Our results suggest that depending on the level of trust to the ML operator and the volume of the available data, distributed training can be an effective way of training a personal information identifier in a privacy-preserved manner. Research material is available at \footnotesize\url{https://github.com/ratmcu/wikipiifed}.

\end{abstract}

\section{Introduction}

\label{sec:intro}

Extraction of Personally Identifiable Information (PII) from unstructured text is a crucial task in many industries such as healthcare (e.g \cite{li2017anonymizing,kushida2012strategies}) legal documents \cite{oksanen2019semantic}, mining of user-generated data \cite{mosallanezhad-etal-2019-deep} and publication process \cite{aura2006scanning}. PII is a laborious task, often necessary for de-identification purposes, among other applications. For example, the extracted information can be used for indexing of documents, categorization and other applications. Identification of PII elements is a laborious task that can be automated by deploying Named Entity Recognition (NER) models \cite{hassan2018anonymization,gralinski2009named}. We formulate PII recognition as a NER task that extracts predefined PII entities. Our goal is to develop NER models to decrease this task's cost by preprocessing documents before manual information extraction.  

Supervised machine learning approaches such as Conditional Random Field (CRF) models, Support vector machines, and extensive feature engineering based on lexical and phrase embeddings have been used to train NER systems \cite{gangLuo,passos2014lexicon,ratinov}. With improvements of deep learning models, recurrent neural networks and specially LSTM models became the default model for training NER systems \cite{chiu2016named}. Recently, a combination of pre-trained transformer-based language models and linear or recurrent prediction layers achieved the state-of-the-art in most NER tasks \cite{dai2019named,fraser2019extracting}.

Training a NER model for extraction of PII demands a massive corpus of text, rich in personal information, which raises privacy concerns in the process of data annotation and model training. Although NER models achieve high performance in cross-validation settings, the generalization of off-the-shelf models remains poor \cite{fu2020rethinking}. For training a robust PII recognizer, a customized domain-specific annotated dataset is needed \cite{chen2015study}. In this work, our goal is to bring privacy to the front line of designing a PII extractor from dataset creation to model training.

We consider a scenario where an institution intends to build an assistant tool to decrease the cost of manual PII extraction. We assume that the institution has accumulated documents alongside their corresponding PII fields over the years, but PII elements are not necessarily marked within the text. This is the typical case for many institutions (such as hospitals and banks), which have manually extracted PII elements for years. For example, a hospital has access to patients' names and ages for every specific health record. However, the locations of occurrences of name and age within the documents are unknown. Also, the name and age can come in various forms and lengths when mentioned in free text. To build a useful training dataset for a PII recognizer, the hospital needs to mark phrases related to name and age in the free text through text mining. However, sharing this data for annotation and training involves privacy considerations. We consider the following steps to ensure our process is compliant with the privacy of data subjects. 

\begin{itemize}
    \item Annotating the free text programmatically  without the need for sharing the data for human annotation.
    \item Distributed storing of annotated documents so that the data can be kept in authorized locations. 
    \item Remote training of the PII extraction model without the need for sharing annotated documents with machine learning practitioners. 
\end{itemize}

To conduct a reproducible research, we show the feasibility of the proposed approach on a dataset collected from Wikipedia and share the created dataset and results with research community. Our contributions are as following:

\begin{itemize}
    \item We create and release an automatically labeled dataset comprised of 77703 sentences from Wikipedia biography pages annotated for 5 classes of personal information.
    \item We develop a method for remote training of a transformer-based model on distributed datasets, using PySyft platform. 
    \item We explore the impact of label noise and dataset size on the performance of remotely trained NER models. 
\end{itemize}

\section{WikiPII Dataset}
\label{sec:data}

Our goal is to create and annotate a customized textual dataset for training a PII extractor (the scenario described in Section \ref{sec:intro}). Approaches like snorkel \cite{ratner2017snorkel} had been embracing noise of automatic annotations and compensated the noise by adding to the volume of inexpensive data.  We took the same approach for annotating Wikipedia pages and benefited from the fact that a version of entities was available in the infobox. We used the infobox to generate noisy and inexpensive data annotations, whereas snorkel uses multiple noisy parallel annotation functions and weak-supervision from alternative sources.

We collected our data from Wikipedia biography pages because 1) they are rich in terms of PII, 2) with the infobox available on each page, they comply with our assumption of having access to extracted PII, 3) they are publicly available and can be shared and used as a benchmark for research purposes. Similar automated annotation tasks such as \citet{nothman2013learning} utilized a broader set of Wikipedia pages and contain general CoNLL style \cite{conll2003shared} (location, organization, person, miscellaneous) classes of entities and does not focus on granular personal information as ours. We refer to this dataset as WikiPII and release this data for further research.   

\subsection{Data Collection }
\label{subsec:datacolelction}

\begin{figure}
\begin{center}
\includegraphics[width = 8cm]{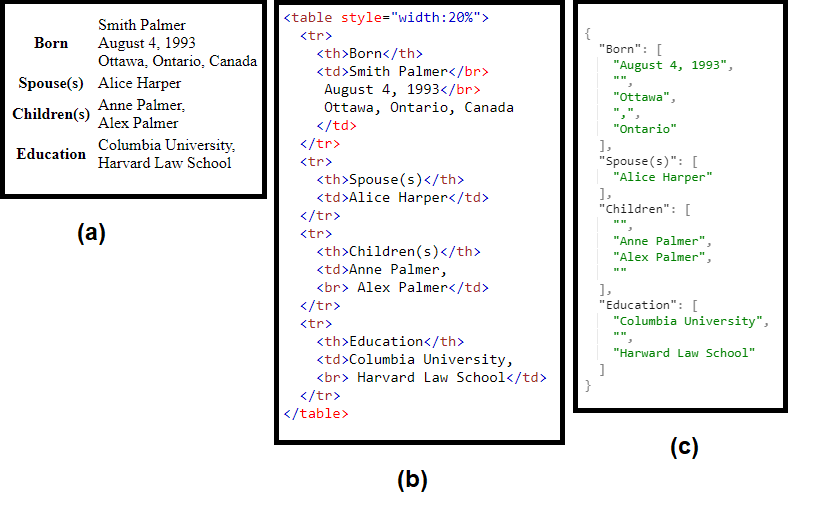}
\end{center}
\caption{Infobox a) viewed on web page, b) in HTML format, c) converted to a dictionary} 
\label{fig:infoboxhtmlandextracted}
\end{figure}

We scraped our raw textual data from biography page entries of living people in Wikipedia (about 900K pages).
For programmatic annotation of each page's textual body, we first read the HTML-coded infobox and converted it to a PII element dictionary, using the BeautifulSoup\footnote{\url{https://www.crummy.com/software/BeautifulSoup/bs4/doc/}} package. An example of this conversion is demonstrated in Figure \ref{fig:infoboxhtmlandextracted}.

Next, we normalized the similar entity types to acquire consistent entity types across all pages. For example, the spouse’s name can come under the titles’ Spouse’,’ Spouse(s)’ and’ Spouses’, which are all normalized to the ‘SP’ tag. After normalization, we manually inspected the entities and chose the ones with high coverage in the dataset. At last, we decided to include BD (date of birth), PR (names of parents), SP (names of spouse(s)), CH (names of children) and ED (terms of education institutes attended). Our final tags and their corresponding infobox entries are shown in Table \ref{table:association} \footnote{We first extracted birthplace from the infobox, but places are mentioned in several formats such as town, province, country, etc. We omit this entity in the final tagging.}.

\begin{table}[ht!]
\centering
\begin{tabular}{p{2cm}|p{4.5cm}} 

 \footnotesize{\textbf{Tag }} & \footnotesize{\textbf{Corresponding entries in infobox}}  \\ [0.5ex] 
 \hline
 \hline
\footnotesize{Birth Date (BD)} &\footnotesize{'Born', 'Born:'}  \\
 \hline
\footnotesize{Parents (PR)} &\footnotesize{'Parent', 'Parent(s)', 'Parents', 'Father', 'Father’s name', 'Mother', 'Mother’s name}\\
 \hline
\footnotesize{Spouses (SP)}     &	\footnotesize{'Spouse', 'Spouse(s)', 'Spouses' }\\
 \hline
\footnotesize{Children (CH)}   &	\footnotesize{'Children' }\\
 \hline
\footnotesize{Education (ED)}   &	\footnotesize{'Education', 'High school', 'High school:', 'Law School', 'School', 'Schools', 'College', 'College(s)', 'Colleges', 'Alma mater','Almat mater' }\\[1ex]

\end{tabular}
\caption{Tags included in our dataset and corresponding entities in infobox.}
\label{table:association}

\end{table}

 \begin{figure*}
   \centering
   \includegraphics[width = 16cm]{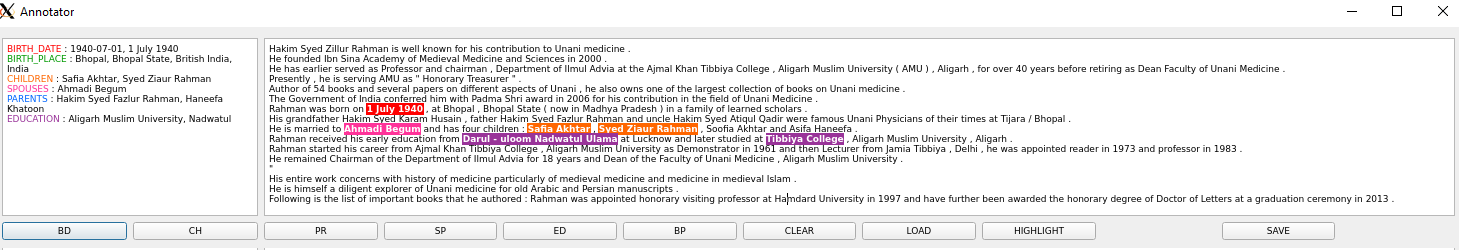}
   \caption{Annotator UI for manual annotation.} 
    \label{fig:annotuiafter}
 \end{figure*}

\subsection{Labeling of Entities in Text}

Once the PII was extracted from the infobox, we had to locate them in the text and generate a tag for each word to create an annotated dataset. This step's main challenge is that the mentions of entities in the text might be variations of the ones extracted from the infobox. 

We parsed the textual body into sentences and removed citation brackets and numerals using regular expressions. For each tag shown in Table \ref{table:association}, we develop a function that takes the extracted phrase from infobox and locates that phrase within the free text. We combine two different methods of matching to get a more accurate match. Our entities are subcategories of places, organizations, persons and dates, which SpaCy already covers. We leveraged the part-of-speech and named entity recognition capabilities of the  SpaCy\footnote{We used the $en\_core\_web\_lg$ pipeline from \url{https://spacy.io/usage/facts-figures##benchmarks}, which is highly accurate in NER task and optimized in terms of speed.} package to find noun chunks, person names, locations, organizations and dates. Then, depending on the type of the entity, we choose a subset of extracted phrases and use fuzzy string matching\footnote{We utilized the implementation published at \url{https://github.com/axiak/fuzzyset/} for fuzzy matching.} to find the closest phrase to our target phrase. For example, for the tag, ED (education), we extract all the organization names by SpaCy. We then use fuzzy matching to find the variations of the education institute pulled from the infobox.

\begin{figure*}[h]
   \centering
\begin{tabular}{p{5cm}p{10cm}}
\includegraphics[width = 5cm]{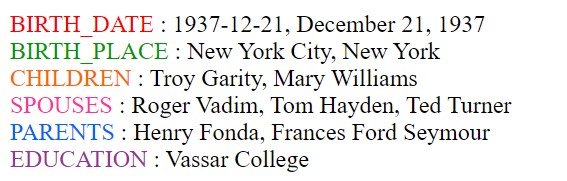}&\includegraphics[width = 10cm]{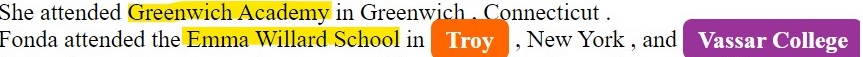} \\
\includegraphics[width = 5cm]{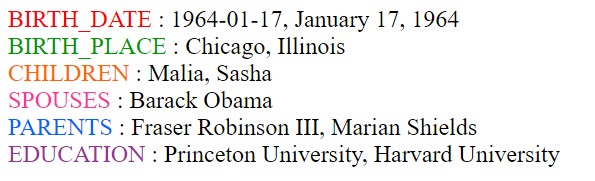}&
\includegraphics[width = 10cm]{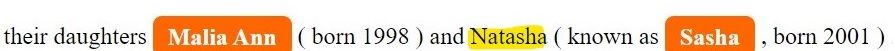}\\ 
\includegraphics[width = 5cm]{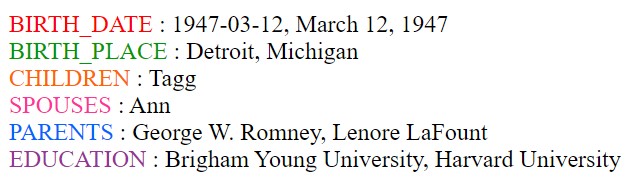} &\includegraphics[width = 10cm]{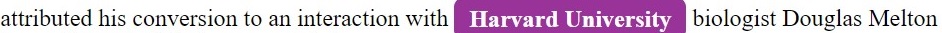}\\
\end{tabular}
\caption{Examples of mistakes in automated annotation.  }
 \label{fig:annot-mistakes} 
\end{figure*}

\begin{table*}[h]
\centering
\small
\begin{tabular}{p{4cm}p{1cm}p{1.5cm}p{1cm}p{1cm}p{1cm}p{1cm}p{1cm}} 
 \hline
 Data split / annotation method     &   Pages  & Sentences & BD & PR & SP  & CH & ED \\ [0.5ex]
 \hline
 training/automatic    & 20039      & 77703 & 16883 & 6326  & 25163 & 10824 &	24365 \\ 
  validation/automatic   & 2744       & 12267 & 2512  & 1509  & 3844  & 1846  &	3831  \\  
 test/automatic        & 307        & 2051  & 303   & 331   & 609   & 604   &	534   \\
 test/manual & 91         & 320   &  76   & 50    & 80    & 62    &    92    \\ 
[1ex] 
 \hline
\end{tabular}
\caption{Count of pages, sentences which contain at least one of the target entities and number of mentions per each class of entities for different splits of the WikiPII dataset. }
\label{table:entitycount}
\end{table*}

We used the BIO scheme for tagging of words. NER is a sequence to sequence learning task that predicts a label for each word, specifying whether the word is within or outside an entity and the entities' type. In the BIO format, the tags `B\_', `I\_', and `O\_'  mark the beginning, inside and outside of an entity, respectively. For example, `B\_CH' specifies the beginning of a phrase tagged as `Children'. The combination of these tags specifies the boundary and tag of the extracted entity. Therefore, error analysis of an NER task is based on errors in tag and boundary detection. These error are reflected in the evaluation metrics described in Section \ref{sec:eval}.

\subsection{Manual Annotation}

To evaluate the quality of the programmatic annotation, we manually annotated a subset of the pages. We selected pages that include highest numbers of entities and made sure that the manually annotated dataset contains 50 to 100 mentions of each class. Manual annotation is done by re-annotating the entities already found by the automated annotator. A human annotator can choose to confirm, reject or correct the labels created by the automatic annotation. We designed a user interface for the manual annotation where the annotator had access to the infobox elements and their corresponding tags. An example of the designed annotator user interface is shown in Figure  \ref{fig:annotuiafter}.

Figure \ref{fig:annot-mistakes} shows examples of common mistakes in the automatic annotation. Entities extracted from the infobox are shown in the left column. The yellow entities are missed by automated annotation and corrected by the human annotator. These entities are missed because they are missing from the infobox. Also, since the automatic annotation does not consider the context, it cannot resolve ambiguities. In the example of Figure  \ref{fig:annot-mistakes} `Troy' is a city name but is tagged as CH (children) since it appears as a child name in the infobox. Also, `Harvard University', which is tagged as an education institute, is not a PII element for the main subject of the page but an affiliation of someone else. In manual annotation, `Troy' and `Harvard University' will be corrected and not tagged as an entity.

\subsection{Statistics of WikiPII dataset}

Our data source contains over 900K entries. Our annotation method could only use a little over 23K entries due to formatting changes in the Wikipedia pages where infobox is not available. We filtered the sentences that do not include any of our target entities. The dataset contains a large number of PII instances belonging to over 23K individuals worldwide belonging to 5 classes. 
Separate splits of the created dataset and numbers of entities contained are presented in Table \ref{table:entitycount}.

\section{Evaluation of PII extraction}
\label{sec:eval}

Averaged F-score is a common metric to evaluate a NER system. However, the definition of a True Positive and a True Negative prediction is not always trivial. Since identifying an entity involves finding both the span and type of the entity, some of the system's predictions can be partially correct. Multiple evaluation schemes have been developed. Shared tasks such as IREX \cite{sekine1999irex}, and CoNLL \cite{conll2003shared} only gave credit to the exacted entities with the exact type and boundary matches. Other works have adopted \textit{type matching} or \textit{partial matching} evaluation schemes, which reward partially correct entity extractions \cite{tsai2006various,chinchor-sundheim-1993-muc,segura2013semeval}. Learning-based evaluation methods are developed to predict the user experience in specific tasks \cite{bionlpisar}.

\begin{table*}[ht]
\centering
\begin{tabular}{|c|c|c|c|c|c| } 
 \hline
 \textbf{\footnotesize{Predicted Entity}}& \textbf{\footnotesize{strict}}& \textbf{\footnotesize{exact}}  & \textbf{\footnotesize{type}}&\textbf{\footnotesize{partial}} & \textbf{\footnotesize{implication in PII extraction}} \\ 
 \hline
 \footnotesize{\color{red}{$_{name}$}\color{black}\colorbox{BurntOrange}{Adam London.} } &\cmark \cmark &\cmark \cmark&\cmark \cmark&\cmark \cmark& \footnotesize{no need for correction.}\\
 \hline
\footnotesize{Adam \color{red}{$_{name}$}\color{black}\colorbox{BurntOrange}{London.} }&\xmark &\xmark&\cmark&\cmark& \footnotesize{boundary should be corrected.}\\
 \hline
 \footnotesize{\color{red}{$_{place}$}\color{black}\colorbox{BurntOrange}{Adam London.} }&\xmark &\cmark\cmark&\xmark&\cmark& \footnotesize{type should be corrected.} \\
 \hline
 \footnotesize{Adam \color{red}{$_{place}$}\color{black}\colorbox{BurntOrange}{London.} }&\xmark &\xmark&\xmark&\cmark& \footnotesize{entity is located, boundary and type should be corrected.} \\
 \hline
\end{tabular}
\caption{Examples of predicted entity with respect to various evaluation metrics. \xmark indicates no reward, \cmark indicates half point reward and \cmark \cmark indicates a full reward. }
\label{tab:metricsner}
\end{table*}

PII extraction is a sensitive task, and a fully automatic system cannot be reliable. Instead, the output of such systems are used to augment the performance of manual PII extraction. 
In practice, when a human is in the loop, partial matching can reduce the manual effort of PII extraction. We adopted the metrics introduced by the MUC-5 task \cite{chinchor-sundheim-1993-muc}, and SemEval-13 task 9 \cite{segura2013semeval} and implemented the following evaluation metrics ordered in terms of strictness: 

\begin{itemize}
    \item \textbf{Strict Matching:} rewards a prediction only if boundary and type of entity match with gold standard label. This metric evaluates the system in a fully automated PII extraction setting.
    \item \textbf{Exact Boundary:} rewards a prediction if the boundary of extracted entity matches the gold standard labeling. This metric evaluates the system where the human annotator relies on boundaries predicted by the system and only corrects the label if necessary.  
    \item \textbf{Type Matching:} rewards the strict matches and partially ($\times$0.5) rewards the extracted entities where the type is correct and boundary overlaps with the gold standard. This metric evaluates the system where the human annotator relies on types predicted by the system and only corrects the boundary if necessary.
    \item \textbf{Partial Boundary:} rewards strict matches and partially ($\times$0.5) rewards where the boundary overlaps with the gold standard label regardless of type. This metric evaluates the system where the human annotator relies on the location of predictions  and corrects both label and boundary if necessary.
\end{itemize}   

Table \ref{tab:metricsner} shows examples of predicted entities by a PII recognizer with respect to the evaluation metrics and the cost of correction in a human-in-the-loop PII recognition task.

\section{PII Extraction model}

First, we evaluate the automated annotation compared to the manual annotation.  Then we use the automatically annotated train set to train a BERT-based NER model with a fully connected linear layer as the prediction layer. We then evaluate the performance of the trained PII recognizer on both automatically and manually annotated test sets.

\subsection{Comparison between Manual and Automatic Annotation}

To evaluate the automatic annotation, we take manual annotations as the gold standard and score the corresponding automated annotations with the metrics described in Section \ref{sec:eval}. Table \ref{tab:manualvsauto} shows the results of this evaluation. As discussed in Section \ref{sec:eval}, we used different metrics to evaluate this model based on the real-application scenario. For example, partial metric evaluates the scenario where the model is used to assist the human annotator in locating the entities. We observe that the rule-based annotation tool leads to high levels of noise. With partial evaluation, we conclude that automatic annotation spots about half of the entities correctly, but the boundary and type might not be fully correct. On the other side, the strict metric indicates that about one-third of the entities are perfectly annotated. Also, the type metric is higher than the exact metric, indicating that automatic annotation performs better in predicting types than boundaries. This is expected because of the complexities and subjectivity of boundary identification.

\begin{table}[ht!]
\small{
\centering
\begin{tabular}{ccccc}
\cline{2-5}

           &	strict  &	exact & type    &	partial \\
\hline

precision     &    0.31	&  0.32      &	0.39&	0.46 \\
recall        &	0.45	&  0.46     &	0.57 &	0.65 \\
F1-score          &	0.37	&  0.38     &	0.47  &	0.54 \\ 
\hline
\end{tabular}
\caption{Evaluation of automated annotation compared to the manual annotation.}
\label{tab:manualvsauto}
}
\end{table}

\subsection{Performance of PII Extraction Model}
\label{subsec:central-performance}
We fine-tuned a BERT-based NER model with the training split of the automatically annotated WikiPII dataset to build a PII recognizer and tested the trained model with the test split of automatically annotated dataset and the manually annotated test set. We choose a batch size of 128 sentences and a maximum length of 50 tokens and present the results for one epoch of training. The optimum number of epochs varies between 1 and 3 for different datasets, but for the sake of comparison we choose to run all experiments with one epoch. Table \ref{tab:testscoremanual-automated} shows the results of this experiment. 

We observed that despite the high level of noise in the automatically annotated training dataset the trained NER model reaches an acceptable performance. This is due to the large size of the automatically annotated dataset. As \citet{rolnick2017deep} showed, deep learning models are robust to label noise when the size of the dataset is adequately large. We observed that the partial metric is 80\%, which indicates a significant decrease in manual cost of PII extraction.  While these predictions might still need corrections of type and boundary the system can locate most of the entities. From the strict metric, we conclude that half of the PII elements are predicted correctly in label and span and do not need any correction. Comparing of the exact and type metric shows that in most cases the system predicts the label correctly and boundaries need to be corrected.

\begin{table*}[ht!]
\centering
\small{
\begin{tabular}{ccccc||cccc} 
\cline{2-9}
&\multicolumn{4}{c}{Automatically annotated } & \multicolumn{4}{c}{ Manually annotated} \\
\cline{2-9}
&strict &exact & 	type&partial  &strict &exact& type&partial\\
\hline

 precision& 0.64 & 0.72 & 0.70 & 0.74 & 0.55 & 0.56& 0.68 & 0.79  \\ 
 recall & 0.62 & 0.69& 0.68 & 0.72& 0.56 & 0.56 & 0.68& 0.80  \\

 F1-score & 0.64 & 0.70& 0.69 & 0.73& 0.55 & 0.56  & 0.68 & 0.80  \\
\hline

\end{tabular}
\caption{Test accuracy of the BERT-based NER model on both test sets}
\label{tab:testscoremanual-automated}
}
\end{table*}

\section{Distributed Training }

Modern deep learning models are known as data-hungry algorithms. In the task of PII extraction, sharing data across organizations will lead to more robust models. However, sharing of personal data in a central location involves concerns of privacy. To mitigate the risk of data breaches, we can train machine learning models in a distributed fashion while leaving the data in a location governed by the data owners. In this work, we explore Federated Learning (FL) \cite{yang2019federated} for training a NER model with noisy labelled data. Federated learning involves training statistical models over remote data centers, such as mobile phones or hospitals, while keeping data localized without requiring transfer of the whole dataset to a central location. 

To implement FL, we use the PySyft framework, developed by \textit{OpenMined} \footnote{\url{http://www.openmined.org/}}. 
 This framework is developed in PyTorch and provides the platform for executing tensor operations remotely \cite{ryffel2018generic}. PySyft has been developed under the theme \textit{"Answer questions you cannot see"}, to perform machine learning inference with zero knowledge about the specifics of the data.

In this framework, a central entity orchestrates the training scenario. Data is maintained and tagged by its owners at a remote location. At each data location, a worker follows the commands of the central entity. The model is transferred to the remote location, and updates are completed remotely at each training iteration. Subsequently, the final model is updated by averaging weights, averaging remote gradient updates or consecutive updates at each dataset location \cite{li2020federated}.

 \subsection{Federated Training of BERT-based Model}
 
BERT (Bidirectional Encoder Representations from Transformers) is a transformer-based language model pre-trained on a massive corpus of written text. BERT and a series of language models belonging to BERT's family form the backbone of today's deep learning NLP models. The language model generates a vector representation for the input text and passes it on to the downstream task. The pre-trained language model is usually fine-tuned with the task-specific data during task-specific learning. In this work, we only use BERT-based NER model, with a fully connected linear layer as the prediction layer, but the general idea applies to other transformer-based language models.Training and testing splits are the same as the ones used in Section \ref{subsec:central-performance}. 

The input to a BERT model contains three tensors: Token type id (specifies single sentence or double sentence use of the model), position ids (specifies the position of the token in the sentence), and input ids (specifies the id of the word in the vocabulary) \cite{devlin2018bert}. Except for input ids, all the other inputs are tensors generated dynamically at the training time. The pre-trained tokenizer model generates these inputs to be based on the dimensions of the input sentence batches. 
 
PySyft is designed in a way that abstracts the remote tensor objects by wrapping them around an empty tensor located centrally.  Wrapping the tensor is the process of maintaining an empty local tensor object, while executing tensor operations on the remote tensor through the network. This method abstracts the location separation and allows the central worker to operate on tensor objects just as they were situated centrally. One drawback of this wrapping-based abstraction is that the functions such as size querying are operated on the local empty tensor rather than the native real tensor located in the remote worker.  For that reason, the PySyft framework cannot query the dimensions of the input data tensors while operating in a remote worker \cite{ryffel2018generic}.

  For remote tokenization through PySyft, we modified the model to carry these inputs as static non-trainable parameters embedded in the form of tensor buffers. Using PySyft, we can move a model between the remote workers and the central worker using the API calls. Initially, these APIs were developed to handle the trainable parameters of the models among workers involved in federated learning. We contributed to the PySyft framework's codebase by developing a federated BERT tokenizer method,  which handles the movement of non-trainable parameters and allows full remote functionality of the model. Our implementation of the \textit{BERT-base} model for remote operation will be released for further research.  

 \subsection{Training Scenarios}
 We deploy two settings of FL to share training data in a privacy-preserved manner. In practice one of these scenarios might be preferred depending on how much trusted the central worker is.   

 \begin{itemize}
     \item federated/central: A trusted central operator can receive data batches from remote data holders
     \item federated/remote: A mistrusted central operator sends model to a remote data holders 
\end{itemize}

\begin{figure}[ht]
\centering
         
\begin{subfigure}[b]{0.4\textwidth}
\centering
\includegraphics[width=6cm]{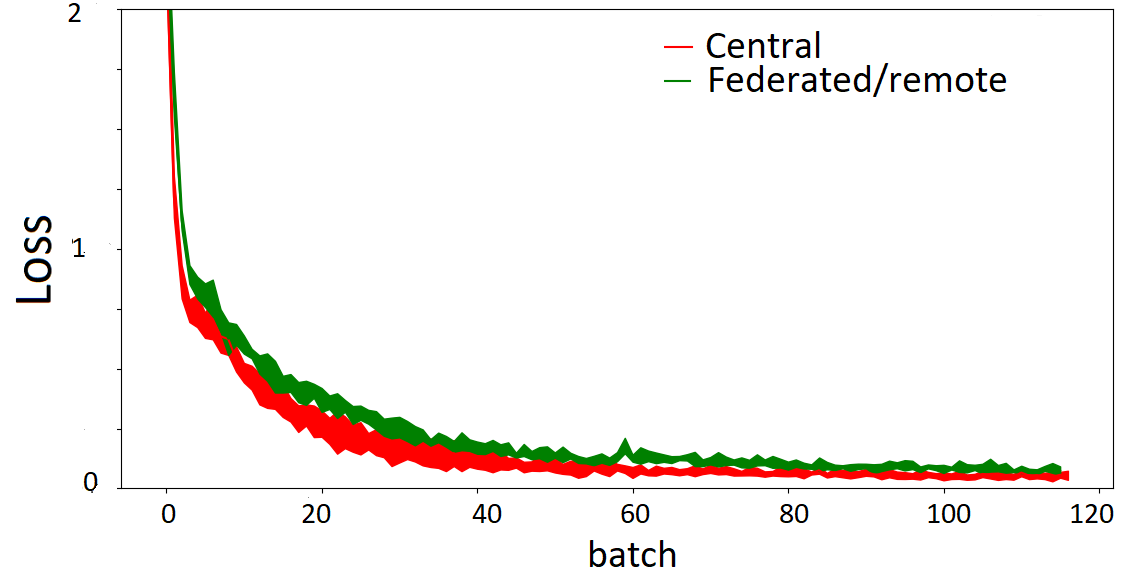}
 \caption{}
\label{fig:fed1}
\end{subfigure}
\hfill
\begin{subfigure}[b]{0.4\textwidth}
\centering
\includegraphics[width=6cm]{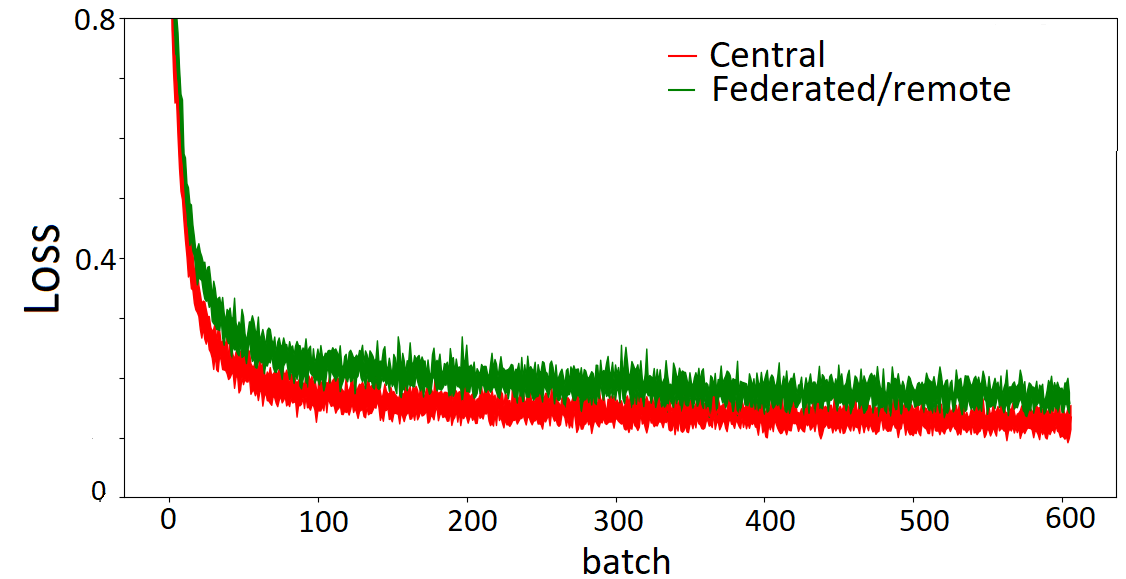}
 \caption{}
\label{fig:fed3}
\end{subfigure}
\caption{Training loss for central vs federated/remote on a) CoNLL-2003 and b) WikiPII dataset }
        \label{fig:three graphs}
\end{figure}

In scenario 1, the operator is trusted to receive data from remote sources and updates the model in the central location. Data batches from distributed sources are called using the federated training iterator. Received data batches contribute to forward pass and back-propagation operations. Then the operator discards the data batch as agreed. Here the operator has full control over the data batches, and the BERT tokenizer works in its typical mode. The only different operations with respect to central training are data transfers from the remote workers towards the central worker. These transfers might lead to information loss because of data compression. Also, from machine learning perspective, distributed data cannot be shuffled randomly and data batches might be imbalanced which has an impact on the final performance of the model.

In scenario 2, the operator is not fully trusted, so the batch of data can not be fully transferred to the central operator. Federated training iterator holds the locations of the data holders or remote workers. The model owned by the central operator is sent to the remote worker and allowed to be remotely executed. Only the central operator's commands are allowed to reach the remote worker guaranteeing the central operator is not breaching into the data. In this scenario, we use our remote tokenization method. Interactions for this training involve sending the model, sending commands to execute the model, and receiving the trained model parameters back. Model weights are received back by the central operator after training for all the batches of data belonging to the remote worker. Then the model is sent to the other workers. An epoch is completed when the model cycles all the workers.

\subsection{Performance of Model Trained with Distributed Data and Noisy Labels}

To gain insight into the impact of distributed training on NER models' performance, besides the WikiPII dataset, we trained our model on the widely used NER dataset, CoNLL-2003. Similar to our central training (Section \ref{subsec:central-performance}), we restrict our experiments to one epoch of training. We simulated both scenarios with only two virtual workers and recorded the training loss to investigate how remote learning impacts the model's convergence. Also, for simplicity, we assume the remote workers' availability at all the times a central worker requests their computational resources for training, which might not be the real-world scenario and will require planning and robustness.

We observed that the case of federated/central training does not impact the convergence of the model. Figures \ref{fig:fed1} and \ref{fig:fed3} show the convergence of the loss when model trained on two workers in federated/remote scenario compared to the typical centralized training, for both CoNLL2003 and WikiPII. In the case of CoNLL2003, where the annotations are of gold-standard quality, the federated training does not significantly impact the model's convergence. In WikiPII, with noisy labels, federated/remote training leads to higher loss function values. However, this impact is not detrimental.   

\begin{table}[t]
\centering
\small{
\begin{tabular}{lcc} 
 \hline
\textbf{ Dataset/Setting} & no. of workers  & F1 score \\ 
 \hline

  \textbf{CoNLL2003} &&\\
  $\ \ \ $ central & N/A & 0.90 $\pm 0.005 $ \\ 
 $\ \ \ $ federated/remote & 2  & 0.85 $\pm 0.003 $ \\ 
 $\ \ \ $ federated/central & 2 & 0.90 $\pm 0.008 $  \\ 
 \hline

 \textbf{WikiPII} &&\\
 $\ \ \ $ central & N/A  & 0.70 $\pm 0.006 $ \\ 
 $\ \ \ $ federated/remote & 2 &  0.56 $\pm 0.02 $ \\ 
 $\ \ \ $ federated/central & 2 &  0.70 $\pm 0.01 $ \\ 
[1ex] 
\hline

\end{tabular}
\caption{Exact F1-score for central vs federated model }
\label{tab:finalscore}
}
\end{table}

The exact F-scores trained under our FL implementations are summarized in Table \ref{tab:finalscore}. We observed very close final model performance between the federated learning with centrally operated and typical centralized training. Federated training with the mistrusted central operator deviates from central training, with higher loss convergence values and a reduced final performance score, for both CoNLL2003 and WikiPII datasets. This observation can be explained by the loss of information in weight compression while transferring the model. 

\subsection{Effect of Dataset Size}

\begin{figure}[ht]
\centering
\includegraphics[width=7.5cm]{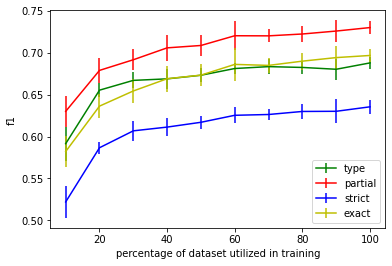}
\caption{F1-scores vs. the training dataset size as more workers participate in federated/central training}
\label{fig:datsize}
\end{figure}

Federated learning is most useful where multiple data holders participate in the training process. In reality, different distributed sources contributing to training can carry imbalanced amounts of data and features, which can have a negative impact on the results.  Here we measure the effect of increasing the dataset size by increasing the number of workers.  We randomly divided the training dataset among ten workers and, starting from 2 workers, increased the number of workers participating in the training process. Figure \ref{fig:datsize} shows the change of different types of F-score as more workers are utilized, and the dataset size increases as a result. We used the federated/central scenario here, which was shown to achieve comparable performance to central training. To control for the random sampling, we repeat each experiment 10 times and average the acquired F1-score. The error plot in Figure \ref{fig:datsize} demonstrates this experiment's final results for all the metrics. 

In general, we observed that when the size of the noisy annotated data increases, higher performances are achieved. Since automatic labelling of data is inexpensive, generating and sharing noisy labelled data is a promising way of achieving high-quality models. However, note that the standard deviation of F-scores can be considerable. This observation indicates that the imbalances of distributed data can drastically impact the final model.  

\section{Discussion}

In this work, our goal is to use the historical data accumulated in an organization to build a customized NER for a human-in-the-loop PII recognition tool. We expect that this tool can significantly decrease the cost of manual extraction by locating PII entities in the free text. 

First, we assume that the organization has access to a corpus of unstructured documents along with the structured dataset containing their corresponding PII entities. We propose that these parallel datasets can be used to create a noisy annotated training set. Our method of automatic annotation is based on matching of phrases and the raw data is not exposed to a third party for annotation. Using Wikipedia biography pages as an example, we show the feasibility of creating a noisy annotated dataset and training a PII recognition model in a privacy-preserved fashion. Our automatic annotation is inexpensive. Therefore it can generate large volumes of annotated datasets to compensate for the label noise. This is in line with previous work showing that deep learning is robust against noise when trained with massive noisy datasets \cite{rolnick2017deep}. 

Furthermore, we looked at the feasibility of distributed training in cases that multiple organizations have similar datasets and are willing to collaborate to build more robust models but cannot share the data due to privacy concerns. We showed that where the operator is trusted, distributed training will not affect performance regardless of the annotation quality. For both CoNLL2003 (clean annotations) and our WikiPII dataset (noisy annotations), the F-score of NER models does not suffer from distributed training. However, when the operator is not trusted, the F-score is impacted and the drop of F-score is more significant in the case of the noisy dataset. In model transfer, all the parameter tensors of the model go through the simplification, serialization and compression steps followed by decompression, de-serialization and decompression steps. We suspect this mechanism affects the precision of the weights. In future work, a rigorous analysis should be carried out to analyze the effect of object transfers in a distributed system. 

Lastly, in the federated/central scenario, we showed that the increase in the dataset size is a promising way to achieve higher accuracies. Distributed training allows organizations to share their data which results in a bigger size of the data. We conclude that there is a trade-off between the drop in performance because of the distributed training and the increase in performance because of the higher volume of data.  

This work has limitations. NER is a very challenging task, and it is difficult to achieve a fully reliable NER model for a sensitive task such as PII extraction. Also, even highly accurate NER models can be vulnerable to adversarial attacks \cite{zhang2020adversarial}. For this reason, throughout this work, we only envisioned this system to assist human annotators by locating the entities and suggest a highly likely tag. Although this system does not reach very high performance, it is still instrumental in reducing the cost of PII extraction when compared to a fully manual procedure. We only considered a BERT-based NER model, but the general idea applies to other transformer-based NER models. In future, an ensemble of different techniques should be considered to improve the utility of the system. 


\section{Conclusion}
We propose an inexpensive and privacy-preserved method that automatically annotates parallel structured/unstructured datasets to train a customized NER models. The final models can be used to decrease the cost of manual extraction of PII elements by preprocessing the documents in a human-in-the-loop setting. Our results demonstrate that federated training is a promising tool to compensate for label noise by increasing the volume of the noisy labeled dataset.

\section*{Acknowledgement}
We would like to thank IMRSV Data Labs for partial funding  and support in manual annotation. We also acknowledge Mitacs for partially funding this project.

\bibliographystyle{acl_natbib}
\bibliography{naacl2021}
\end{document}